\documentclass[runningheads]{llncs}

\usepackage{eccv}

\usepackage{eccvabbrv}

\usepackage{graphicx}
\usepackage{booktabs}
\usepackage{multirow}
\usepackage{multicol}
\usepackage{colortbl}
\usepackage[normalem]{ulem}

\usepackage[accsupp]{axessibility}  

\usepackage{hyperref}

\usepackage{orcidlink}

\begin{document}

\title{Recurrent Cross-View Object Geo-Localization} 

\titlerunning{Recurrent Cross-View Object Geo-Localization}

\author{Xiaohan Zhang\inst{1,2}\orcidlink{0000-0002-1420-1909} \and 
Si-Yuan Cao\inst{1,3}\thanks{Corresponding author.}\orcidlink{0000-0001-5143-4501} \and
Xiaokai Bai\inst{2}\orcidlink{0009-0002-8382-2976} \and 
Yiming Li\inst{2}\orcidlink{0009-0008-3212-1030} \and
\\ Zhangkai Shen\inst{2}\orcidlink{0009-0002-7996-9086} \and 
Zhe Wu\inst{2}\orcidlink{0009-0009-8846-980X} \and 
Lun Luo\inst{2}\orcidlink{00000-0002-0531-9171} \and 
Qi Ming\inst{4}\orcidlink{0000-0001-7596-171X} \and 
Xiaoxi Hu\inst{5}\orcidlink{0000-0001-5702-1281} \and \\
Hui-Liang Shen\inst{2}\orcidlink{0000-0001-8469-019X}}

\authorrunning{X.~Zhang, et al.}

\institute{Ningbo Global Innovation Center, Zhejiang University, Ningbo 315100, China 
\and
College of Information Science and Electronic Engineering, Zhejiang University, Hangzhou 310027, China 
\and
Jinhua Institute of Zhejiang University, Jinhua 321000, China 
\and
College of Computer Science, Beijing University of Technology, Beijing 100124, China
\and
State Key Laboratory of Intelligent Green Vehicle and Mobility, Tsinghua University, Beijing 100084, China
\\
\email{\{zhangxh2023, cao\_siyuan\}@zju.edu.cn}}

\maketitle

\begin{abstract}
  Cross-view object geo-localization (CVOGL) aims to determine the location of a specific object in high-resolution satellite imagery given a query image with a point prompt. Existing approaches treat CVOGL as a one-shot detection process, directly regressing object locations from cross-view information aggregation, but they are vulnerable to feature noise and lack mechanisms for error correction. In this paper, we propose \textbf{ReCOT}, a \textbf{Re}current \textbf{C}ross-view \textbf{O}bject geo-localization \textbf{T}ransformer, which models CVOGL as a recurrent localization process. ReCOT introduces a set of learnable tokens that encode task-specific intent from the query image and prompt embeddings, and iteratively attend to the reference features to refine the predicted location. To enhance this recurrent process, we incorporate two complementary modules: (1) a SAM-based knowledge distillation strategy that transfers segmentation priors from the Segment Anything Model (SAM) to provide clearer semantic guidance without additional inference cost, and (2) a Reference Feature Enhancement Module (RFEM) that introduces hierarchical attention to emphasize object-relevant regions in the reference features. Extensive experiments on CVOGL benchmarks demonstrate that ReCOT achieves state-of-the-art (SOTA) performance while significantly reducing parameters compared to previous SOTA approaches. Our code is available at \href{https://github.com/Temperature-ai/ReCOT.git}{https://github.com/Temperature-ai/ReCOT.git}. 
\end{abstract}

\section{Introduction}
\label{sec:intro}

Cross-view object geo-localization (CVOGL) aims to determine the geographic location of a specific object indicated by point prompts in a query image on the reference image \cite{DetGeo}. The query images can be captured from devices like phones, autonomous vehicles, robots, and drones, while the reference images are typically high-resolution satellite images. CVOGL is widely used in various applications, such as smart city management \cite{smart_city_1}, disaster monitoring \cite{disater_1, disater_2}, and robot navigation \cite{robot, robot2}. However, the view gap poses challenges \cite{DetGeo}.

Recent cross-view image geo-localization (CVIGL) works \cite{CVM-Net, SAFA, TransGeo, L2LTR, RK-Net} have demonstrated their superiority in handling view gaps. However, CVIGL approaches are fundamentally designed for camera-level localization using retrieval-based approaches \cite{SampleGeo, GeoDTR_plus, SAFA} or fine-grained approaches \cite{OriNet, HC-Net}. In contrast, CVOGL aims to localize specific objects (\emph{e.g.}, a building with a red roof) captured in the query image, which demands prompt-guided and object-aware prediction. Therefore, in CVOGL scenarios, CVIGL approaches can only provide a nearby location for the indicated object \cite{DetGeo}, which is insufficient for precise object-level localization.

\begin{figure}[!htb]
  \centering
  \includegraphics[width=0.8\linewidth]{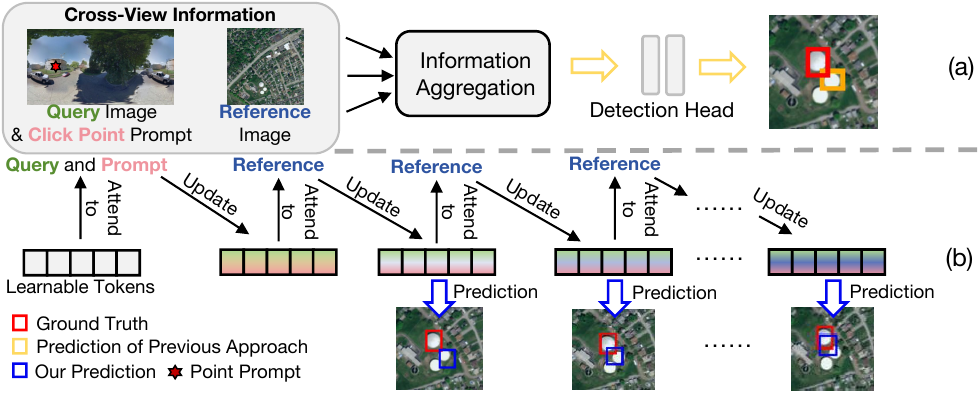}
   \caption{Comparison between the framework of previous CVOGL approaches and ours. (a) Previous approaches treat the CVOGL as a prompt-based detection process, where the model directly regresses the object location based on information aggregation once. (b) Our framework models the CVOGL as a recurrent localization process, where the model iteratively refines the localization through a set of learnable tokens.}
   \label{fig: head}
\end{figure}

To address this, CVOGL approaches emerge recently. Existing approaches \cite{DetGeo, GeoFormer, VaGeo, TROGeo, OCGNet, AttenGeo} typically treat CVOGL as a one-shot detection paradigm, where the model directly regresses the object location based on prompt-guided information aggregation, as shown in Fig. \ref{fig: head}(a). For example, the recent state-of-the-art (SOTA) approach \cite{OCGNet} aggregates information from cross-view images and prompts to produce a spatial attention matrix, which is used to enhance the reference image features. The enhanced features are then fed into several convolutional layers to regress the object location. While efficient and architecturally simple, such a framework lacks a correction mechanism for early-stage prediction errors, making it vulnerable to noise in features \cite{RHWF, DetGeo}. Specifically, since large view gaps yield multiple satellite candidates, the one-shot model must commit to a single spatial hypothesis in one pass, and this early commitment conditions the subsequent cross-view localization \cite{DetGeo}, making errors uncorrectable. 

To cope with this, we propose a \textbf{Re}current \textbf{C}ross-view \textbf{O}bject geo-localization \textbf{T}ransformer (ReCOT). Motivated by the success of iterative refinement strategies \cite{ASTR, IHN, RHWF}, ReCOT models the CVOGL task as a recurrent localization process, as shown in Fig. \ref{fig: head}(b). This serves as the main difference between previous approaches \cite{DetGeo, GeoFormer, VaGeo, TROGeo, OCGNet, AttenGeo} and our framework. Specifically, ReCOT initializes a set of learnable tokens, which interact with the query image feature and prompt embeddings to extract task-specific intent. These tokens then act as recurrent “questioners” that iteratively attend to the enhanced reference image features, progressively extracting object-relevant information and refining the prediction. The proposed recurrent strategy enables our ReCOT to effectively enhance the performance of CVOGL by iteratively refining the initial prediction, as shown in Fig. \ref{fig: head}(b). Compared to one-shot frameworks, recurrent refinement relaxes the one-pass commitment by treating localization as a revisable prediction state, \emph{i.e.}, the model can iteratively revise its spatial hypothesis by re-weighting competing candidates and accumulating object-consistent evidence. Therefore, our recurrent mechanism effectively lifts the performance ceiling, offering greater potential to address more complex scenarios.

Nevertheless, the success of this recurrent strategy in our token-driven framework relies on the semantic clarity of the prompts \cite{VaGeo} and the quality of reference features \cite{RAFT}. Specifically, the learnable tokens are first guided by prompt semantics to extract object-relevant intent, then iteratively attend to the reference features to refine the object location in each recurrent step. Therefore, if prompts lack clear semantic intent or reference features are cluttered, the tokens may not accumulate correct task-specific cues across iterations, leading to suboptimal performance. To tackle this, we introduce two complementary methods: (1) a SAM-based knowledge distillation strategy, which injects prior knowledge from a large-scale model into the prompt embeddings to boost prompt understanding while avoiding computational cost during inference, and (2) a Reference Feature Enhancement Module (RFEM), which emphasizes object-relevant reference features through hierarchical attention. These components provide clean visual and semantic cues, enabling the tokens to effectively accumulate task-specific information during iterative refinement.

We evaluate ReCOT on CVOGL benchmarks \cite{DetGeo, OCGNet}. It achieves state-of-the-art (SOTA) performance while significantly reducing parameter count compared to previous one-shot approaches \cite{DetGeo, GeoFormer}, and runs at a competitive inference speed. In summary, our contributions are as follows:

\begin{itemize}
    \item We propose ReCOT, a novel framework for CVOGL that models the task as a recurrent localization process, where learnable tokens iteratively attend to reference features to refine object localization.
    \item We introduce a SAM-based knowledge distillation strategy that transfers prior knowledge from a large foundation model into the prompt embeddings, providing clearer semantic guidance without adding inference cost.
    \item We design the RFEM, which leverages the proposed hierarchical attention to highlight object-relevant regions in the reference feature, thereby facilitating the recurrent localization process.
\end{itemize}

\section{Related Work}
\label{sec: Related Work}

\textbf{Cross-View Image Geo-Localization (CVIGL).} 
CVIGL is an image-level geo-localization task. It aims to identify the geographic location of the camera by finding the most correlated reference image \cite{SampleGeo, GeoDTR_plus, SAFA} or the most correlated position \cite{OriNet, HC-Net, SliceMatch} on the reference image. 
Previous CVIGL works can be roughly divided into direct metric learning-based methods and geometry-based methods. The former \cite{CVIGL1, CVIGL2, CVIGL3, CVIGL4, CVIGL5, CVIGL6, CVIGL7, CVM-Net, L2LTR, RK-Net, TransGeo} directly learns viewpoint-invariant features without considering geometric configurations. In contrast, the geometry-based approaches \cite{CVIGL10, CVIGL11, CVIGL12, CVIGL8, CVIGL9, SAFA} explicitly reduce the viewpoint discrepancy between ground- and aerial-view images by using geometric configurations such as orientation information. However, existing CVIGL methods determine the approximate location of the query image on the reference image, thus being ineffective for object-level localization \cite{DetGeo}.

\textbf{Cross-View Object Geo-Localization (CVOGL).} CVOGL is an object-level geo-localization task aiming to determine the geographic location of an object indicated by a point prompt in a query image on the reference image. Considering the vital role of CVOGL and the limitations of current CVIGL methods, DetGeo \cite{DetGeo} first proposes the CVOGL task and a corresponding method based on the spatial correlation of cross-view images and an object detection framework \cite{YOLOv3}. As a pioneering work, DetGeo encodes the point prompt into a Euclidean-distance-based positional map and fuses it with the query image features via element-wise addition. The fused features are then pooled to obtain a channel-only query vector, which interacts with the reference image features through matrix multiplication to produce a spatial attention map. This map is applied to the reference features to highlight the object region, and the enhanced reference features are finally passed through a convolution-based detection head to regress the object location. Subsequent works \cite{VaGeo, TROGeo, OCGNet, AttenGeo} enhance cross-view feature aggregation and prompt embedding based on the framework of DetGeo \cite{DetGeo}. Despite progress, existing CVOGL methods still rely on one-shot detection, which is sensitive to noisy features \cite{IHN} and lacks mechanisms for error correction. In contrast, we model CVOGL as a recurrent localization process and propose ReCOT to address these limitations.

\section{Method}
\label{sec: Methodology}

Fig. \ref{fig: overall} presents the architecture of ReCOT, which models CVOGL as a recurrent localization process. A set of learnable tokens is initialized to encode task-specific intent from the query image features and prompt embeddings. These tokens act as recurrent “questioners” that iteratively extract object-relevant cues from the reference features and refine the predicted location through cross-attention mechanisms. Additionally, we introduce two complementary methods to enhance this recurrent process: (1) a SAM-based knowledge distillation strategy, which transfers segmentation priors from the Segment Anything Model (SAM) to enhance prompt semantics, and (2) a Reference Feature Enhancement Module (RFEM), which provides object-relevant reference features through the proposed hierarchical attention for the recurrent stage.

\subsection{Recurrent Localization Framework}
\textbf{Motivation.} As shown in Fig. \ref{fig: head}, existing CVOGL approaches follow a one-shot detection paradigm, directly predicting the object location from enhanced reference features. However, such frameworks are often sensitive to feature noise and lack a mechanism for error correction \cite{IHN, RHWF}. Fundamentally, CVOGL can be regarded as a cross-view matching problem, where recurrent strategies have shown superior robustness across domains \cite{IHN, RAFT}. Inspired by this, we model CVOGL as a recurrent localization process. Moreover, unlike dense matching tasks \cite{RHWF, ASTR, roma}, CVOGL is prompt-driven and focuses on object-level semantic matching. This calls for a representation that can both encode semantic intent and drive iterative refinement. To this end, we draw inspiration from the class token in vision transformers (ViT) \cite{vit}, and introduce a set of learnable tokens that absorb task-specific semantics from the query and prompt. Acting as semantic carriers, these tokens can recurrently interact with the reference feature to enable step-wise prediction.

\begin{figure*}[!htb]
  \centering
  \includegraphics[width=0.9\linewidth]{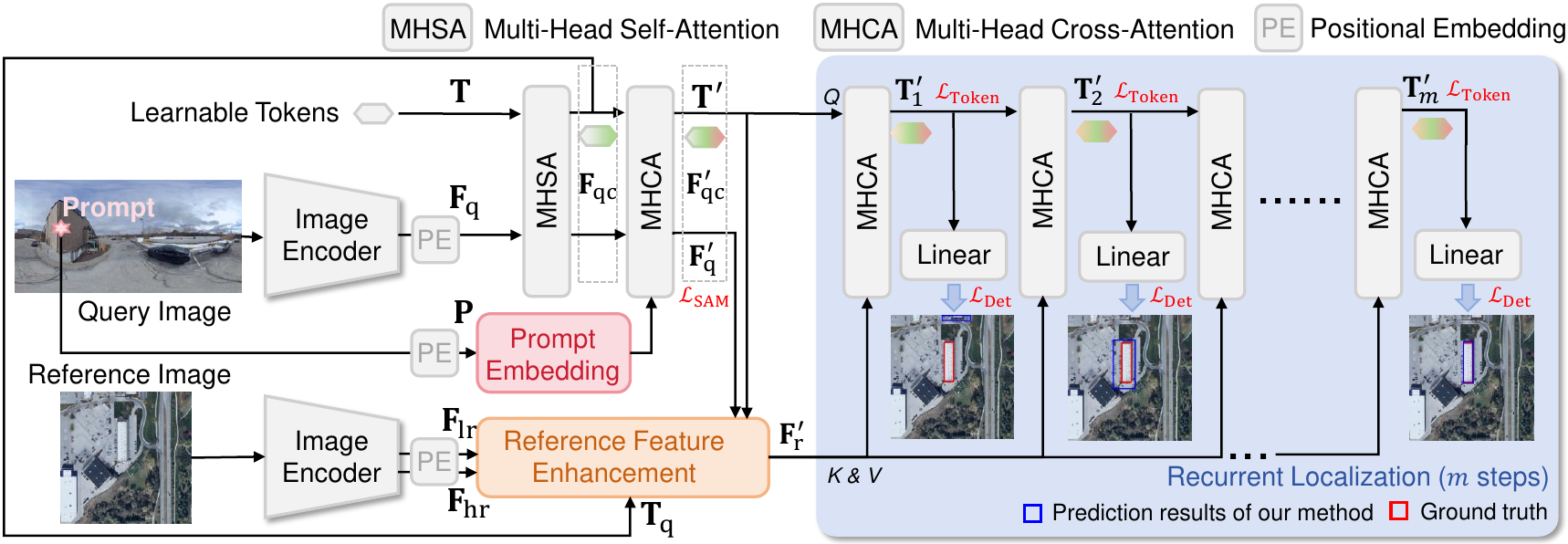}
   \caption{Architecture of our recurrent cross-view object geo-localization transformer (ReCOT). ReCOT models the CVOGL task as a recurrent localization process, which leverages a set of learnable tokens to extract information from cross-view images and prompts to recurrently refine the prediction. Notably, all ``MHCA" and ``Linear" components in the recurrent steps share the same weights, following the typical design of recurrent architectures. We perform the positional embedding following DETR \cite{DETR}}
   \label{fig: overall}
\end{figure*}

\textbf{Structure.} We initialize a set of learnable tokens $\textbf{T} \in \mathbb{R}^{n \times c}$, where $n$ and $c$ denote the number of tokens and the feature dimension, respectively. To enable $\textbf{T}$ to extract object-relevant information from the reference features, $\textbf{T}$ needs to first acquire task-specific semantics from the query image feature $\textbf{F}_{\text{q}} \in \mathbb{R}^{h_{\text{q}}w_{\text{q}} \times c}$ and the point prompt embedding $\textbf{P} \in \mathbb{R}^{c}$. Here, $h_{\text{q}}$ and $w_{\text{q}}$ denote the height and width of the query feature, respectively. Specifically, we concatenate $\textbf{T}$ with $\textbf{F}_{\text{q}}$ along the spatial dimension, yielding $\mathbf{F}_{\text{qc}} \in \mathbb{R}^{(n + h_{\text{q}}w_{\text{q}}) \times c}$. Following standard operations in Vision Transformers (ViT) \cite{vit}, we apply self-attention to $\textbf{F}_{\text{qc}}$, allowing $\textbf{T}$ to aggregate global context from the query image. The resulting tokens are denoted as $\textbf{T}_{\text{q}}$. To further inject object-level intent, the embedded point prompt $\textbf{P}$ interacts with $\textbf{F}_{\text{qc}}$ through cross-attention. This enables the prompt to semantically guide the tokens, allowing $\textbf{T}_{\text{q}}$ to further incorporate object-level context. After interaction, we denote the concatenated feature and tokens as $\textbf{F}'_{\text{qc}}$ and $\textbf{T}'$, respectively. The $\textbf{F}'_{\text{qc}}$ and $\textbf{T}_{\text{q}}$ are further utilized to enhance reference features in RFEM. The enhanced reference feature is denoted as $\textbf{F}'_{\text{r}} \in \mathbb{R}^{h_{\text{r}}w_{\text{r}} \times c}$, where $h_{\text{r}}$ and $w_{\text{r}}$ denote the height and width of the reference feature, respectively.

After acquiring task-specific semantics from the query and prompt, we use $\textbf{T}'$ to perform recurrent localization, as shown in Fig. \ref{fig: overall}. Let $\textbf{T}'_{i}$ denote the token state at the $i$-th refinement step, where $i \in [0,1,2,\dots,m]$. At each step, $\textbf{T}'_{i}$ attends to the enhanced reference feature $\textbf{F}'_{\text{r}}$ to extract object-relevant cues and update the task-specific intent. Formally, the update process is defined as
\begin{equation}
    \textbf{T}'_{i+1} = \text{MHCA}(\textbf{T}'_{i}, \textbf{F}'_{\text{r}})
\end{equation}
where $\text{MHCA}(\cdot, \cdot)$ denotes the multi-head cross-attention module. Here, $\textbf{T}'_{i}$ serves as the query, while $\textbf{F}'_{\text{r}}$ acts as the key and value. This recurrent attention mechanism enables iterative refinement, where the tokens $\textbf{T}'$ progressively accumulate task-specific semantics and extract increasingly relevant information from the fixed reference feature $\textbf{F}'_{\text{r}}$. The $\textbf{T}'_{i}$ of each step is fed into a linear layer to predict an updated object location, allowing the model to gradually converge toward a precise localization.

In each refinement step $i$, we introduce a loss $\mathcal{L}_{\text{Token}}$ to guide the generation of $\textbf{T}'_{i}$. Specifically, since $\textbf{T}'_{i}$ is expected to contain the task-specific intent in cross-view features, it should be able to highlight the required object area on $\textbf{F}'_{\text{r}}$. Therefore, in each refinement step, we first aggregate $\textbf{T}'_{i}$ along the spatial dimension to generate a global embedding $\textbf{T}''_{i}$, and use $\textbf{T}''_{i} \in \mathbb{R}^{1 \times c}$ and $\textbf{F}'_{\text{r}}$ to produce an aggregation map $\widehat{\textbf{m}}_{\mathrm{o}} \in \mathbb{R}^{1 \times h_{\text{r}} \times w_{\text{r}}}$. This can be expressed as
\begin{equation}
    \textbf{T}''_{i}=\text{Sum}(\textbf{T}'_{i} \cdot \text{Softmax}(\textbf{T}'_{i}))\text{,}
\label{aggregation T}
\end{equation}
\begin{equation}
    \widehat{\textbf{m}}_{\mathrm{o}i} = \sigma({\textbf{T}''_{i}}\textbf{F}'^{\mathsf{T}}_{\text{r}})\text{,}
\end{equation}
where $\sigma(\cdot)$ and $\text{Softmax}(\cdot)$ denote the sigmoid and softmax functions, respectively. $\text{Sum}(\cdot)$ denotes summation along the spatial dimension. We then utilize a box-level mask $\textbf{m}_{\mathrm{o}i}$ produced using the ground truth box to supervise the generation of $\widehat{\textbf{m}}_{\mathrm{o}}$, which can be expressed as
\begin{equation}
    \mathcal{L}_{\text{Token}_{i}}(\textbf{m}_{\mathrm{o}}, \widehat{\textbf{m}}_{\mathrm{o}i}) = \mathcal{L}_{\text{bce}}(\textbf{m}_{\mathrm{o}}, \widehat{\textbf{m}}_{\mathrm{o}i}) + \mathcal{L}_{\text{dice}}(\textbf{m}_{\mathrm{o}}, \widehat{\textbf{m}}_{\mathrm{o}i})\text{,}
\label{loss}
\end{equation}
where $\mathcal{L}_{\text{bce}}(\cdot, \cdot)$ and $\mathcal{L}_{\text{dice}}(\cdot, \cdot)$ denote the binary cross-entropy loss and the Dice Loss \cite{dice-loss}, respectively.

\begin{figure}[!htb]
  \centering
  \includegraphics[width=0.7\linewidth]{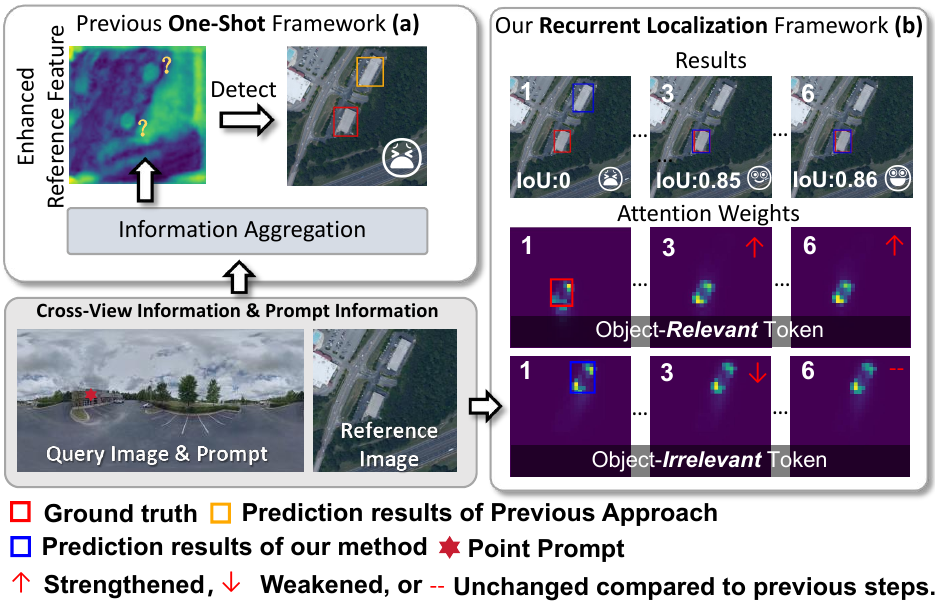}
   \caption{Comparison between the previous one-shot framework and our recurrent localization framework. (a) Previous approaches \cite{DetGeo, VaGeo, TROGeo, OCGNet, AttenGeo} rely on single-shot information aggregation, which is sensitive to noisy enhanced features and often leads to incorrect predictions. (b) Our ReCOT iteratively refines predictions through learnable tokens. The attention weight visualizations show that object-relevant tokens progressively focus and strengthen around the target, while object-irrelevant tokens weaken and stabilize to background patterns.}
   \label{fig: motivation}
\end{figure}

\textbf{How ReCOT works.} As shown in Fig. \ref{fig: motivation}(a), previous one-shot detection CVOGL approaches \cite{DetGeo, VaGeo, TROGeo, OCGNet, AttenGeo} rely on a single forward information aggregation and are thus sensitive to noisy features \cite{IHN, RHWF}. Our ReCOT adopts a recurrent localization mechanism that iteratively refines predictions. The visualization in Fig.~\ref{fig: motivation}(b) of cross-attention weights between tokens and the reference feature reveals the inner dynamics of this refinement process. It can be seen that different tokens focus on different regions of the reference feature, indicating a form of token-level specialization. For object-relevant tokens, their attention gradually concentrates and intensifies around the object region across recurrent steps, reflecting the ability to correct early prediction error and progressively refine the prediction. In contrast, object-irrelevant tokens experience a decrease in their attention responses and eventually stabilize to background patterns once they no longer contribute to the object localization. This behavior highlights the competitive nature of token updates, \emph{i.e.}, multiple tokens initially compete to explain different parts of the reference feature \cite{DETR}, but those correlated with the object receive positive feedback and their attention weights are amplified over recurrent steps, leading to iterative convergence. Such dynamics demonstrate the effectiveness of recurrent refinement mechanism in suppressing irrelevant regions while enhancing object-relevant cues.

\subsection{SAM-based Knowledge Distillation}
\label{subsec: SAM}
\textbf{Motivation.} Point prompt understanding is essential for CVOGL to correctly locate objects. However, point prompt itself suffers from semantic ambiguity, leading to unsatisfactory performance \cite{SAM}. To address this, we propose a SAM-based knowledge distillation strategy to boost the prompt understanding of ReCOT. The incorporation of SAM \cite{SAM} is motivated by an observation that SAM can give a mask with a clear indication of the required object using point prompts and corresponding images. However, directly applying SAM during inference incurs large computation overhead. Hence, we leverage predictions of SAM as supervision signals, transferring its knowledge through knowledge distillation. 

\textbf{Structure.} We extract the query feature $\textbf{F}'_{\text{q}}$ from the previous concatenated feature $\textbf{F}'_{\text{qc}}$. Notably, the $\textbf{F}'_{\text{q}}$ has been interacted with prompt embedding and is supposed to contain the object-level semantics. Therefore, we process it using a lightweight convolutional head followed by a sigmoid activation to generate a segmentation map $\widehat{\textbf{m}}_{\mathrm{q}}$. Meanwhile, we use SAM to generate a pseudo ground-truth mask $\textbf{m}_{\mathrm{SAM}}$ from the query image and point prompt. This mask is used to supervise the segmentation out of $\textbf{F}'_{\text{q}}$ via $\mathcal{L}_{\text{SAM}}$, which can be defined as
\begin{equation}
    \mathcal{L}_{\text{SAM}}(\textbf{m}_{\text{SAM}}, \widehat{\textbf{m}}_{\mathrm{q}}) = \mathcal{L}_{\text{bce}}(\textbf{m}_{\text{SAM}}, \widehat{\textbf{m}}_{\mathrm{q}}) + \mathcal{L}_{\text{dice}}(\textbf{m}_{\text{SAM}}, \widehat{\textbf{m}}_{\mathrm{q}})\text{.}
\label{loss_sam}
\end{equation}

\subsection{Reference Feature Enhancement Module}
\label{subsec: refm}

\textbf{Motivation.} In CVOGL, the reference feature $\textbf{F}_{\text{r}}$ extracted by the backbone encoder is typically generic and background-dominated, lacking object-level specificity before prompt interaction \cite{DetGeo, VaGeo}. In our framework, guiding $\textbf{F}_{\text{r}}$ to focus on the expected object indicated by the prompt can significantly ease the downstream recurrent localization \cite{RHWF, RAFT}. To this end, we propose the Reference Feature Enhancement Module (RFEM), which enhances reference features into a more object-aware representation $\textbf{F}'_{\text{r}}$. Unlike previous approaches \cite{DetGeo, VaGeo, OCGNet, TROGeo} that attempt to clearly highlight the object features through one-shot feature enhancement, RFEM serves as a preparatory module to filter irrelevant features and provide more object-relevant information for subsequent recurrent localization. The core of RFEM lies in its hierarchical attention design. We first perform spatial attention to highlight the query-relevant regions in the reference feature, as the expected object is likely confined to these regions. We then apply cross-attention guided by the query and prompt cues to refine these regions with object-level semantics. This spatial-to-cross hierarchy yields a cleaner and more informative reference feature $\textbf{F}'_{\text{r}}$, which significantly benefits the iterative localization process of ReCOT.

\begin{figure}[!htb]
  \centering
  \includegraphics[width=0.8\linewidth]{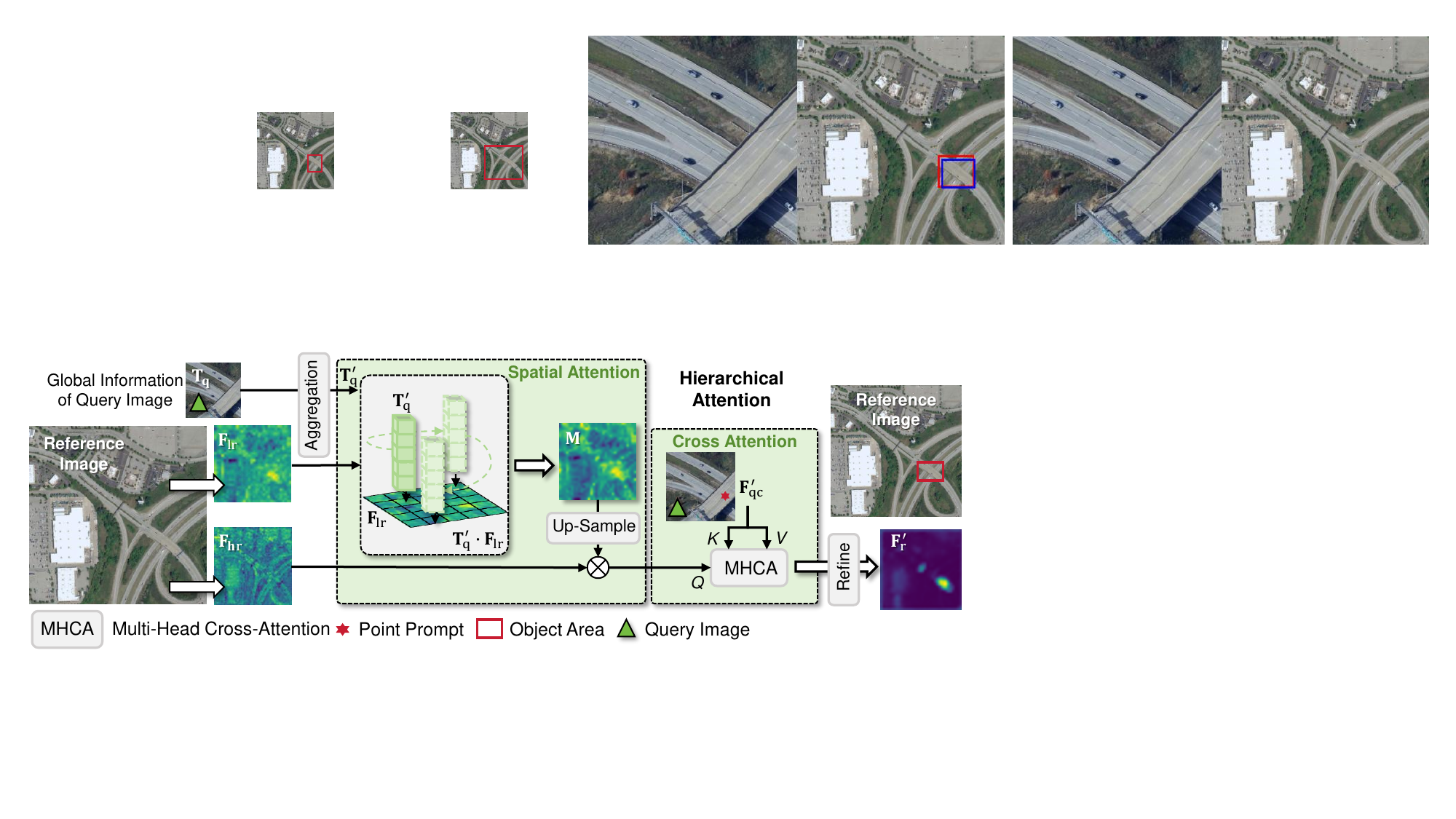}
   \caption{Architecture of reference feature enhancement module (RFEM). RFEM enhances reference features through a hierarchical attention pipeline to obtain $\textbf{F}'_{\text{r}}$.}
   \label{fig: rfem}
\end{figure}

\textbf{Structure.} As illustrated in Fig.~\ref{fig: rfem}, we utilize two levels of reference features, \emph{i.e.}, a low-resolution semantic feature $\textbf{F}_{\text{lr}} \in \mathbb{R}^{h_{\text{r}} \times w_{\text{r}} \times c}$ that captures scene semantics, and a high-resolution detailed feature $\textbf{F}_{\text{hr}} \in \mathbb{R}^{2h_{\text{r}} \times 2w_{\text{r}} \times c}$ that preserves fine-grained spatial structures \cite{MRF3Net}.

\textit{Spatial Attention:} We leverage the spatial attention to highlight query-relevant features. Specifically, we first aggregate the query tokens $\textbf{T}_{\text{q}}$ into a global descriptor $\textbf{T}'_{\text{q}}$ (Eq.~(\ref{aggregation T})). Notably, the $\textbf{T}'_{\text{q}}$ contains only the global information of the query image without prompt guidance. This descriptor correlates with $\textbf{F}_{\text{lr}}$ via a dot-product operation to produce a spatial attention map as 
\begin{equation}
\textbf{M} = \sigma(\textbf{T}'_{\text{q}}\textbf{F}^{\mathsf{T}}_{\text{lr}}),
\end{equation}
which highlights query-relevant regions. The $\sigma(\cdot)$ denotes the sigmoid function. The attention map is then up-sampled and applied to $\textbf{F}_{\text{hr}}$ by element-wise multiplication, narrowing the focus to potential object areas.

\textit{Cross Attention:} We leverage the cross-attention to incorporate detailed query features and prompt semantics for further refining the reference feature. The filtered $\textbf{F}_{\text{hr}}$ is subsequently refined via cross-attention with $\textbf{F}_{\text{qc}}$, which encodes fine-grained prompt cues and detailed query features. This cross-attention step sharpens the object-relevant responses and injects object-level semantics into the reference representation. 

Finally, the updated feature is down-sampled and refined through spatial attention \cite{CBAM} and self-attention \cite{vit} to generate $\textbf{F}'_{\text{r}}$, which is then passed to the recurrent localization stage in ReCOT.

\textbf{Why multi-resolution reference features?} 
The reference image typically contains both global scene information and fine-grained object details. 
The low-resolution semantic reference feature $\mathbf{F}_{\text{lr}}$ extracted from deeper layers of the backbone encodes more scene-level context but loses spatial details due to down-sampling. Therefore, we utilize $\mathbf{F}_{\text{lr}}$ to enhance query-relevant regions. 
Conversely, the high-resolution reference features $\mathbf{F}_{\text{hr}}$ preserve fine structural details but are dominated by background noise. Thus, it needs to be refined by the attention map produced by $\mathbf{F}_{\text{lr}}$, and then RFEM can utilize it to perform object-level enhancement. This design leverages the advantages of multi-resolution features, which are crucial for object-level feature enhancement in CVOGL \cite{OCGNet}.

\subsection{Loss Function}
\label{subsec: loss function}
Our training objective $\mathcal{L}$ is defined as a weighted sum of the localization loss $\mathcal{L}_{\text{local}}$ and the auxiliary loss $\mathcal{L}_{\text{aux}}$. It can be defined as 
\begin{equation}
    \mathcal{L} = \mathcal{L}_{\text{local}} + \alpha\mathcal{L}_{\text{aux}}\text{,}
\end{equation}
where $\mathcal{L}_{\text{local}}$ denotes the sum of DETR-style detection losses $\mathcal{L}_{\text{Det}}$ computed at each recurrent localization step (see Fig.~\ref{fig: overall}). The auxiliary loss $\mathcal{L}_{\text{aux}}$ is the sum of $\mathcal{L}_{\text{Token}_{i}}$ across all recurrent steps and the SAM-based distillation loss $\mathcal{L}_{\text{SAM}}$. The balancing coefficient $\alpha$ is set to 1 in all experiments.

\section{Experiments}
\label{sec: Experiment}

\subsection{Datasets}
\label{sec: Datasets}
\textbf{CVOGL} dataset \cite{DetGeo} divides the task into ``Ground $\rightarrow$ Satellite” task and ``Drone $\rightarrow$ Satellite” task. It contains 6,239 pairs of ``Ground $\rightarrow$ Satellite” view and 6,239 pairs of ``Drone $\rightarrow$ Satellite” view query and reference images. The ground view query images, drone view query images, and satellite view reference images are sized to 512 $\times$ 256, 256 $\times$ 256, and 1024 $\times$ 1024, respectively. Each cross-view task uses 4,343 pairs for training, 923 pairs for validation, and 973 pairs for testing. Our experiments utilize the training set to train the model and select the best model on the validation set for evaluation on the test set.

\textbf{CVOGL-few-shot} \cite{OCGNet} dataset introduces additional object categories and is designed for the ``Drone $\rightarrow$ Satellite” task. It annotates 4 new categories beyond the CVOGL \cite{DetGeo} dataset, \emph{e.g.}, lake, parking, slide, and port. The dataset consists of 28 training samples and 24 test samples, with 7 training instances per category across 4 new classes. This dataset is used only for few-shot experiments.

\subsection{Evaluation Metrics}
\label{sec: Metrics}
Following DetGeo \cite{DetGeo}, we adopt $\text{Acc@}0.25$ and $\text{Acc@}0.50$ for evaluation. The Acc@0.25 and Acc@0.50 measure the prediction accuracy under a specific intersection over union (IoU) threshold $t$ between the predicted box $b_{p}$ and ground-truth box $b_{g}$ as 
\begin{equation}
    \text{Acc@}t = \frac{1}{n}\sum\limits_{i=1}^n\psi(t)\text{,}
\end{equation}
where
\begin{equation}
\psi(t) = \left\{
\begin{array}{lr}
1, \text{IoU}(b_{p}, b_{g}) \ge t\\
0, \text{otherwise} 
\end{array}
\right.\text{,}
\end{equation}

\begin{equation}
    \text{IoU}(b_{p}, b_{g}) = \frac{|b_{p} \cap b_{g}|}{|b_{p} \cup b_{g}|}\text{.}
\end{equation}

For each pair of the query and reference image, we select the box with the highest confidence output by our model as the final prediction box. Additionally, we use the parameter and frames per second (FPS) to show the model efficiency. Higher $\text{Acc@}0.25$, higher $\text{Acc@}0.50$, higher FPS, and fewer parameters denote better performance. 

\subsection{Implementation Details}
\label{sec: Implementation Details}
We conduct all experiments on 4 NVIDIA GeForce RTX 4090 GPUs. For training, we adopt the AdamW \cite{adamw} as the optimizer and set the initial learning rate to $2.5\times10^{-5}$, the weight decay rate to $1\times10^{-4}$, and the batch size to 16. Since CVOGL is a relatively new task, we select some CVIGL approaches \cite{CVM-Net, RK-Net, L2LTR, SAFA, TransGeo, GeoDTR_plus} and representative CVOGL approaches \cite{DetGeo, VaGeo, OCGNet, TROGeo, HALO, GeoFormer, AttenGeo}, as our comparison methods. We produce the results of CVIGL approaches following previous works \cite{DetGeo, VaGeo, OCGNet, TROGeo}. All CVOGL methods are trained to convergence to ensure a fair comparison. Additionally, we adopt Swin Transformer (Swin-t) \cite{swin} as the image encoder for its superior performance in various fields \cite{MENet, DuST, Trinity-Net, Restormer}. During training, we use a larger value $m = 6$ so that the model learns a stable multi-step refinement process, while at inference we use $m$ that is selected on the validation set to balance accuracy and efficiency.

\subsection{Ablation Study}
\label{sec: ablation}

\begin{table}[!htb]
\caption{Ablation study on the recurrent localization steps $m$ of ReCOT in terms of Average IoU (Avg.IoU)$\uparrow$,  $\text{Acc@}0.50(\%)\uparrow$, and FPS $\uparrow$ on the \emph{validation set} of CVOGL dataset. We test the inference speed using one NVIDIA GeForce RTX 4090 GPU. \textbf{Bold} and \underline{Underline} indicate the best and second-best results, respectively. We set $m=5$ for optimal results.}
\scriptsize
\centering
\setlength{\tabcolsep}{0.5mm} 
\renewcommand\arraystretch{1.2} 

\begin{tabular}{{c|ccc|ccc}}
\hline
\toprule
\multirow{2}{*}{Steps $m$} & \multicolumn{3}{c|}{Ground$\rightarrow$Satellite} & \multicolumn{3}{c}{Drone$\rightarrow$Satellite} \\ \cline{2-7}
                          & Avg.IoU          & Acc@0.50          & FPS & Avg.IoU & \multicolumn{1}{c}{Acc@0.50}  & FPS                        \\ \midrule
1             & 0.354  & 42.25   & 26.9 & 0.553    & \multicolumn{1}{c}{67.06} & 27.2                   \\
2             & 0.359   & 42.90   & 26.2 & 0.554    & \multicolumn{1}{c}{67.28} & 26.5                    \\
3             & \textbf{0.365}   & \underline{43.55}   & 25.6 & 0.555 & \multicolumn{1}{c}{67.17} & 25.7                   \\
4             &  \underline{0.362}   & 43.34   & 24.8  & 0.555    & \multicolumn{1}{c}{\underline{67.61}} & 24.9                   \\
\rowcolor{gray!10} 5             & \textbf{0.365}    & \textbf{43.66}   & 24.1  &  \textbf{0.557}   & \multicolumn{1}{c}{\textbf{67.82}}  & 24.3                  \\
6             &  \underline{0.362}    & 43.34   & 23.4  &  \underline{0.556}    & \multicolumn{1}{c}{\underline{67.61}}  & 23.5                  \\
\hline\toprule
\end{tabular}

\label{tabel_ablation_recurrent_stage}
\end{table}

\textbf{How to determine $m$ on the dataset?}
Since recurrent localization constitutes the core mechanism of ReCOT, selecting an appropriate number of refinement steps $m$ is crucial for exploiting its capability.
We determine $m$ following a principled and practically applicable strategy that aligns with common practices of existing iterative refinement approaches \cite{RAFT, FlowFormer, IHN}. Specifically, $m$ can be chosen by monitoring the performance gain per step on the \emph{validation set} and stopping when the gain becomes negligible. Following this strategy, we investigate the impact of $m$ on ReCOT using the validation set of CVOGL dataset \cite{DetGeo}. Increasing $m$ from 1 (one-shot prediction) to 5 yields consistent improvements, with the relatively optimal results achieved at $m{=}5$. Further increasing $m$ beyond the optimal point does not bring additional gains and may slightly degrade performance, which can be attributed to over-refinement and overfitting in deeper iterations \cite{hur2019iterativeresidualrefinementjoint, RHWF, ASTR}. Therefore, unless otherwise specified, we set $m{=}5$ and keep this setting in other reported experiments.

\textbf{Explanation of performance saturation with excessive steps.}
The recurrent mechanism in ReCOT progressively refines localization by predicting residual corrections. As shown in Table \ref{tabel_ablation_recurrent_stage}, performance improves and then gradually saturates with minor fluctuations. In early iterations, residuals correct coarse geometric misalignments, yielding noticeable gains. Once the prediction becomes sufficiently accurate, subsequent residuals mainly capture subtle perturbations, leading to diminishing returns, as observed in other iterative refinement methods \cite{IHN, RHWF}. From an optimization perspective, when the objective is already near a local optimum, further iterations bring marginal improvements while increasing computational cost \cite{RAFT}. Therefore, we set $m=5$ as a balanced trade-off between accuracy and efficiency.

\begin{table}[!htb]
\caption{Ablation study on the components of ReCOT ($m{=}5$) in terms of $\text{Acc@}0.25(\%)\uparrow$ and $\text{Acc@}0.50(\%)\uparrow$ on the \emph{test set} of CVOGL dataset.} 
\scriptsize
\centering
\setlength{\tabcolsep}{0.5mm} 
\renewcommand\arraystretch{1.2} 
\begin{tabular}{{c|cc|cc}}
\hline
\toprule
\multirow{2}{*}{Component} & \multicolumn{2}{c|}{Ground$\rightarrow$Satellite} & \multicolumn{2}{c}{Drone$\rightarrow$Satellite} \\ \cline{2-5}
                          & Acc@0.25          & Acc@0.50          & Acc@0.25 & Acc@0.50                         \\ \midrule
w/o RFEM             & 46.45             & 42.24             & 50.15    & 46.04                     \\ 
w/o $\mathcal{L}_{\text{SAM}}$           & 49.74             &  44.81    & 70.91    & 65.78                         \\
w/o $\mathcal{L}_{\text{Token}}$           & 50.57             & 47.58     & 72.05    & 66.80                         \\
\midrule
\rowcolor{gray!10} ReCOT (Ours)              &  52.00                 &  48.10                & 77.60         & 72.05                                                \\ \hline\toprule
\end{tabular}

\label{tabel_ablation_components}
\end{table}

\textbf{Effect of Complementary Components in ReCOT.} Table \ref{tabel_ablation_components} presents the ablation study of the complementary components in ReCOT on the CVOGL test set. Removing the RFEM module (w/o RFEM) leads to a noticeable performance drop, confirming that early reference feature enhancement is critical for guiding tokens to focus on relevant regions. Similarly, removing the SAM-based distillation loss $\mathcal{L}_{\text{SAM}}$ (w/o $\mathcal{L}_{\text{SAM}}$) results in performance degradation, indicating the importance of accurate prompt semantic understanding. The full ReCOT model achieves the best performance across both scenarios. In addition, removing the token-guidance loss $\mathcal{L}_{\text{Token}}$ (w/o $\mathcal{L}_{\text{Token}}$) also degrades performance, which highlights its role in encouraging the learnable tokens to accurately capture object-relevant areas during recurrent refinement. These results collectively validate that RFEM, $\mathcal{L}_{\text{SAM}}$, and $\mathcal{L}_{\text{Token}}$ complement each other, contributing to robust performance of ReCOT.

\begin{table}[!htb]
\caption{Ablation study inside the RFEM ($m$=5) in terms of $\text{Acc@}0.25(\%)\uparrow$ and $\text{Acc@}0.50(\%)\uparrow$ on the \emph{test set} of CVOGL dataset.} 
\scriptsize
\centering
\setlength{\tabcolsep}{0.5mm} 
\renewcommand\arraystretch{1.2} 

\begin{tabular}{{c|cc|cc}}
\hline
\toprule
\multirow{2}{*}{RFEM} & \multicolumn{2}{c|}{Ground$\rightarrow$Satellite} & \multicolumn{2}{c}{Drone$\rightarrow$Satellite} \\ \cline{2-5}
                          & Acc@0.25          & Acc@0.50          & Acc@0.25 & Acc@0.50                         \\ \midrule
w/o $\textbf{M}$             & 51.18             &  47.48            & 75.44    & 70.30                     \\ 
Replace $\textbf{F}_{\text{lr}}$  with $\textbf{F}_{\text{hr}}$           & 48.00             & 44.19     & 75.85    & 70.09                         \\
Replace $\textbf{F}_{\text{hr}}$  with $\textbf{F}_{\text{lr}}$           & 47.49             & 43.28     &75.64     & 67.11                         \\
\midrule
\rowcolor{gray!10} ReCOT (Ours)              &  52.00                 &  48.10                & 77.60         & 72.05                                                \\ \hline\toprule
\end{tabular}

\label{tabel_ablation_RFEM}
\end{table}

\textbf{Effect of Components Within the RFEM.} Table \ref{tabel_ablation_RFEM} investigates the contributions of components within the RFEM. Removing the weight matrix $\textbf{M}$ (w/o $\textbf{M}$) leads to a noticeable performance drop. This confirms that the spatial attention stage benefits the reference feature enhancement. Furthermore, replacing the low-resolution semantic feature $\textbf{F}_{\text{lr}}$ with the high-resolution feature $\textbf{F}_{\text{hr}}$ and replacing $\textbf{F}_{\text{hr}}$ with $\textbf{F}_{\text{lr}}$ both result in performance degradation, highlighting the importance of leveraging multi-resolution reference features in RFEM. 

Please refer to the supplementary material for more ablation study results on hyper-parameters.

\subsection{Comparison Results}
\label{sec: comparion}

\begin{table}[!htb]
\caption{Comparisons in terms of $\text{Acc@}0.25(\%)\uparrow$, $\text{Acc@}0.50(\%)\uparrow$, Parameter (M) $\downarrow$, and FPS $\uparrow$ on the CVOGL dataset. Following Table \ref{tabel_ablation_recurrent_stage}, we set $m{=}5$ for final results. We test the inference speed on the Drone$\rightarrow$Satellite task using one NVIDIA GeForce RTX 4090 GPU. \textbf{Bold} and \underline{Underline} indicate the best and second-best results, respectively. }
\scriptsize
\centering
\setlength{\tabcolsep}{0.5mm}
\renewcommand\arraystretch{1.2}
\begin{tabular}{c|cc|cc|cc}
\hline
\toprule
 & \multicolumn{2}{c|}{Ground$\rightarrow$Satellite} & \multicolumn{2}{c|}{Drone$\rightarrow$Satellite} \\ \cline{2-5}
\multirow{-2}{*}{Method} &
\multicolumn{2}{c|}{\emph{Test Set}} &
\multicolumn{2}{c|}{\emph{Test Set}} &  & \\ 
 & Acc@0.25 & Acc@0.50 & Acc@0.25 & Acc@0.50 & \multirow{-4}{*}{Param} & \multirow{-4}{*}{FPS} \\ \midrule
CVM-Net \cite{CVM-Net}   & 4.73  & 0.51   & 20.14 & 3.29  & -- & --  \\
RK-Net \cite{RK-Net}     & 7.40  & 0.82   & 19.22 & 2.67  & -- & --  \\
L2LTR \cite{L2LTR}       & 10.69 & 2.16  & 38.95 & 6.27   & -- & --  \\
Polar-SAFA \cite{SAFA}   & 20.66 & 3.19   & 37.41 & 6.58   & -- & --  \\
TransGeo \cite{TransGeo} & 21.17 & 2.88   & 35.05 & 6.47   & -- & --  \\
SAFA \cite{SAFA}         & 22.20 & 3.08   & 37.41 & 6.58   & -- & --  \\
GeoDTR+ \cite{GeoDTR_plus} & 14.19 & 5.14 & 16.03 & 4.73 & -- & --  \\ \midrule
DetGeo \cite{DetGeo}     & 45.43 & 42.24 & 61.97 & 57.66 & \underline{73.8}  & \textbf{29.5} \\
VAGeo \cite{VaGeo}       & 48.21 & 45.22 & 66.19 & 61.87 & --  & -- \\
TROGeo \cite{TROGeo}     & \underline{51.08} & \underline{46.56} & 76.16 & 68.96 & 71.3  & 13.5 \\ 
GeoFormer \cite{GeoFormer}  & 49.54 & 45.32 & 73.79 & 68.96 & 739.1  & 11.1 \\ 
AttenGeo \cite{AttenGeo}     & 50.57 & 46.15 & 70.71 & 62.08 & --  & -- \\
\midrule
ReCOT ($m{=}1$, one-shot) & 49.74   & 46.25 & \textbf{78.31} & \underline{71.74} &\textbf{29.9} & \underline{27.2} \\
\rowcolor{gray!10} ReCOT (Ours) & \textbf{52.00} & \textbf{48.10} & \underline{77.60} & \textbf{72.05} & \textbf{29.9} & 24.3 \\ \hline
\toprule
\end{tabular}

\label{table_accuracy_comparison}
\end{table}

\textbf{Quantitative Results.} Table~\ref{table_accuracy_comparison} compares ReCOT with previous approaches on CVOGL. ReCOT consistently outperforms all competitors, achieving SOTA results on almost all test metrics. These gains are obtained with about 60\% fewer parameters, and ReCOT maintains competitive inference speed. Moreover, in scenarios with larger viewpoint gaps (Ground$\rightarrow$Satellite), increasing $m$ from 1 to 5 yields substantial performance gains. This clearly highlights the robustness and potential of our recurrent design in addressing challenging scenarios.

\begin{table}[!htb]
\caption{Few-shot learning performance between DetGeo \cite{DetGeo}, OCGNet \cite{OCGNet}, and our ReCOT in terms of $\text{Acc@}0.25(\%)\uparrow$ and $\text{Acc@}0.50(\%)\uparrow$  on the \emph{test set} of CVOGL-few-shot dataset \cite{OCGNet}. \textbf{Bold} and \underline{Underline} indicate the best and second-best results, respectively.}
\centering
\scriptsize
\setlength{\tabcolsep}{0.5mm} 
\renewcommand\arraystretch{1.2} 
\begin{tabular}{{c|cc}}
\hline
\toprule
\multirow{2}{*}{Method} & \multicolumn{2}{c}{Drone$\rightarrow$Satellite}      \\ \cline{2-3}
                          & Acc@0.25          & Acc@0.50                                   \\ \midrule
DetGeo \cite{DetGeo}                  & 16.67            & 16.67                                       \\
GeoFormer \cite{GeoFormer}            & \underline{45.83}             & \underline{29.17}                                            \\ 
OCGNet \cite{OCGNet}                  & 29.17             & 25.00                                            \\ \midrule
\rowcolor{gray!10} ReCOT($m{=}1$)             & 37.50                  &  25.00 \\
\rowcolor{gray!10} ReCOT($m{=}2$)             & 41.67              &  \underline{29.17}                            \\
\rowcolor{gray!10} ReCOT($m{=}4$)             & \underline{45.83}                  &  \textbf{33.33}                            \\
\rowcolor{gray!10} ReCOT($m{=}6$)             & \textbf{50.00}                  &  \textbf{33.33}                            \\
 \hline
  \toprule
\end{tabular}

\label{tabel_few_shot}
\end{table}

To further assess robustness, we evaluate few-shot learning on the CVOGL-few-shot dataset~\cite{OCGNet} (Table \ref{tabel_few_shot}). Following OCGNet~\cite{OCGNet}, we initialize ReCOT from the pretrained checkpoints in Table~\ref{table_accuracy_comparison} and fine-tune on CVOGL-few-shot. ReCOT clearly surpasses previous CVOGL approaches~\cite{DetGeo, OCGNet} under few-shot conditions, demonstrating the effectiveness of our design. Besides, increasing the number of recurrent steps $m$ leads to a substantial accuracy gain, highlighting the advantage of our recurrent refinement mechanism under more challenging scenarios (few-shot). Additional comparisons of the backbone can be found in the supplementary material.

\begin{figure}[!htb]
  \centering
  \includegraphics[width=1.0\linewidth]{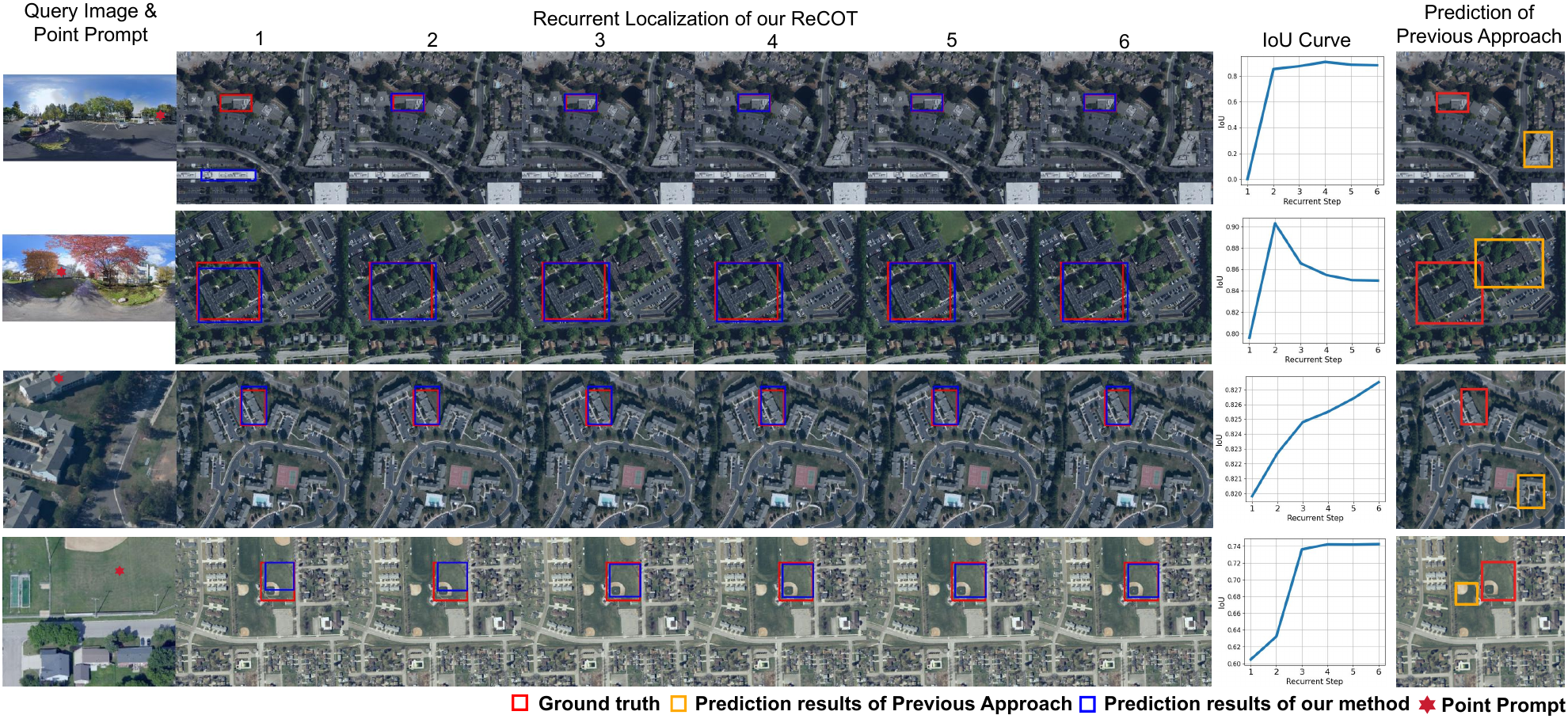}
   \caption{Visualization of some representative results produced by our ReCOT and the representative previous one-shot framework \cite{DetGeo}.}
   \label{fig: result}
   
\end{figure}

\textbf{Qualitative Results.}
Figure~\ref{fig: result} shows representative examples. As recurrent steps proceed, ReCOT progressively refines the bounding boxes and produces more accurate localizations than the single-shot prediction of DetGeo~\cite{DetGeo}. We set $m$ at the dataset level for simplicity and reproducibility. Although the optimal refinement depth may vary per instance, the refinement process consistently converges in practice, demonstrating the stability of our recurrent localization mechanism.   Future work will investigate adaptive stopping criteria for $m$.

\subsection{Discussion}
\textbf{The Role and Core Advantage of the Recurrent Refinement.} Our recurrent localization mechanism fundamentally changes how the model performs CVOGL. Specifically, with $m{=}1$, ReCOT functionally behaves as a \uline{one-shot predictor} performing a single forward prediction without iterative refinement, and achieves performance comparable to existing one-shot approaches in challenging settings (Ground$\rightarrow$Satellite, Acc@0.50: TROGeo \cite{TROGeo} \uline{46.56}, ReCOT($m{=}1$) \uline{46.25}; Few-shot, Acc@0.50: OCGNet \cite{OCGNet} \uline{25.00}, ReCOT($m{=}1$) \uline{25.00}). However, by increasing $m$, ReCOT progressively corrects its initial predictions using the same network and parameters, and gradually achieves the best performance without architectural changes or retraining. This advantage becomes more pronounced in harder few-shot scenarios (Table \ref{tabel_few_shot}), highlighting its robustness. Therefore, our recurrent mechanism effectively lifts the performance ceiling.

Although certain complementary components yield noticeable performance gains (Table \ref{tabel_ablation_components}), their primary role is to enhance the starting point of localization, whereas the recurrent refinement governs the convergence behavior and determines the final attainable accuracy. Our improvements are therefore complementary, with recurrent localization serving as the core driver of robust performance in difficult conditions. This structural capability is absent in prior one-shot architectures \cite{DetGeo, VaGeo, OCGNet, TROGeo, HALO, GeoFormer, AttenGeo}, offering greater potential to address more complex scenarios.

Please refer to the supplementary material for more discussions.

\section{Conclusions}
In this paper, we propose ReCOT, which models the CVOGL as a recurrent localization process. ReCOT introduces learnable tokens to encode task-specific semantics and recurrently refine predictions. We further incorporate a SAM-based knowledge distillation scheme to improve prompt understanding without incurring additional inference costs, and a RFEM to produce object-aware reference features via a hierarchical attention strategy. Extensive experiments on the public benchmarks demonstrate that ReCOT achieves state-of-the-art performance with significantly fewer parameters and competitive inference speed. 

\section*{Acknowledgements}
This work was supported in part by the National Natural Science Foundation of China under Grant 62301484, in part by the Jinhua Science and Technology Bureau Project under Grant 2026-1-022.
%
%

\newpage
\title{Recurrent Cross-View Object Geo-Localization \\ Supplementary Material} 

\titlerunning{Recurrent Cross-View Object Geo-Localization}

\author{Xiaohan Zhang\inst{1,2}\orcidlink{0000-0002-1420-1909} \and 
Si-Yuan Cao\inst{1,3}\thanks{Corresponding author.}\orcidlink{0000-0001-5143-4501} \and
Xiaokai Bai\inst{2}\orcidlink{0009-0002-8382-2976} \and 
Yiming Li\inst{2}\orcidlink{0009-0008-3212-1030} \and \\
Zhangkai Shen\inst{2}\orcidlink{0009-0002-7996-9086} \and 
Zhe Wu\inst{2}\orcidlink{0009-0009-8846-980X} \and 
Lun Luo\inst{2}\orcidlink{00000-0002-0531-9171} \and 
Qi Ming\inst{4}\orcidlink{0000-0001-7596-171X} \and 
Xiaoxi Hu\inst{5}\orcidlink{0000-0001-5702-1281} \and \\
Hui-Liang Shen\inst{2}\orcidlink{0000-0001-8469-019X}}

\authorrunning{X.~Zhang et al.}

\institute{NingBo Global Innovation Center, Zhejiang University, NingBo 315100, China 
\and
College of Information Science \& Electronic Engineering, Zhejiang University, HangZhou 310027, China 
\and
JinHua Institute of Zhejiang University, JinHua, 321000, China 
\and
College of Computer Science, Beijing University of Technology, Beijing 100124, China
\and
State Key Laboratory of Intelligent Green Vehicle and Mobility, Tsinghua University, Beijing 100084, China
\\
\email{\{zhangxh2023, cao\_siyuan\}@zju.edu.cn}}

\maketitle

\section{More Experimental Results}

\subsection{More Ablation Study Results}

\begin{table}[!htb]
\centering
\scriptsize
\setlength{\tabcolsep}{0.5mm} 
\renewcommand\arraystretch{1.2} 
\caption{Ablation study on the hyper-parameter $\alpha$ in terms of $\text{Acc@}0.25(\%)\uparrow$ and $\text{Acc@}0.50(\%)\uparrow$ on the \emph{test set} of CVOGL dataset.} 
\begin{tabular}{{c|cc|cc}}
\hline
\toprule
\multirow{2}{*}{$\alpha$} & \multicolumn{2}{c|}{Ground$\rightarrow$Satellite} & \multicolumn{2}{c}{Drone$\rightarrow$Satellite} \\ \cline{2-5}
                          & Acc@0.25          & Acc@0.50          & Acc@0.25 & Acc@0.50                         \\ \midrule
0.1             & 51.28             & 48.10             & 75.13    & 69.68                     \\ 

\rowcolor{gray!10} 1              &  52.00                 &  48.10                & 77.60         & 72.05                                               \\ 
10           & 48.51             & 45.02     & 76.05    & 70.20                         \\
\hline\toprule
\end{tabular}

\label{tabel_ablation_alpha}
\end{table}

\textbf{Effect of the parameter $\alpha$.}
Table~\ref{tabel_ablation_alpha} shows how the performance of ReCOT varies with different values of the balancing coefficient $\alpha$, which controls the weight between the localization loss $\mathcal{L}_{\text{local}}$ and the auxiliary loss $\mathcal{L}_{\text{aux}}$. We observe that $\alpha = 1$ yields the best performance on both Ground$\rightarrow$Satellite and Drone$\rightarrow$Satellite. A smaller value ($\alpha = 0.1$) weakens the supervision of token alignment and SAM distillation, slightly reducing accuracy. Conversely, a larger value ($\alpha = 10$) overemphasizes auxiliary objectives, causing a notable drop. These results indicate that $\alpha = 1$ achieves the optimal trade-off.

\begin{table}[!htb]
\caption{Ablation study on the hyper-parameter $n$ in terms of $\text{Acc@}0.25(\%)\uparrow$ and $\text{Acc@}0.50(\%)\uparrow$ on the \emph{test set} of CVOGL dataset.} 
\centering
\scriptsize
\setlength{\tabcolsep}{0.5mm} 
\renewcommand\arraystretch{1.2} 

\begin{tabular}{{c|cc|cc}}
\hline
\toprule
\multirow{2}{*}{$n$} & \multicolumn{2}{c|}{Ground$\rightarrow$Satellite} & \multicolumn{2}{c}{Drone$\rightarrow$Satellite} \\ \cline{2-5}
                          & Acc@0.25          & Acc@0.50          & Acc@0.25 & Acc@0.50                         \\ \midrule
1             & 47.89             & 43.37            & 74.31    & 68.86                     \\ 

\rowcolor{gray!10} 100              &  52.00                 &  48.10                & 77.60        & 72.05                                                \\ 
200           & 50.46      & 47.28     &  72.25   & 67.01                        \\
\hline\toprule
\end{tabular}

\label{tabel_ablation_n}
\end{table}

\textbf{Effect of the hyper-parameter $n$.}
Table \ref{tabel_ablation_n} investigates the impact of the number of tokens $n$ on the performance of ReCOT. We observe a clear performance gain when increasing 
$n$ from 1 to 100. This demonstrates that a sufficient number of tokens provides a richer representation of task-specific intent and better coverage of cross-view semantics, which benefits the recurrent refinement process. However, when $n$ is further increased to 200, performance drops across all metrics. This decline is likely due to over-parameterization and token redundancy, which can introduce noise and hinder effective attention learning \cite{vit}, as also observed in token-scaling studies \cite{vit, Deit}. Therefore, we set $n$ to 100 in this work.

\begin{table}[!htb]

\centering
\scriptsize
\setlength{\tabcolsep}{0.5mm} 
\renewcommand\arraystretch{1.2} 

\caption{Ablation study on the type of SAM \cite{SAM} used for $\mathcal{L}_{\text{SAM}}$ in terms of $\text{Acc@}0.25(\%)\uparrow$ and $\text{Acc@}0.50(\%)\uparrow$ on the \emph{test set} of CVOGL dataset.} 

\begin{tabular}{{c|cc|cc}}
\hline
\toprule
\multirow{2}{*}{Type} & \multicolumn{2}{c|}{Ground$\rightarrow$Satellite} & \multicolumn{2}{c}{Drone$\rightarrow$Satellite} \\ \cline{2-5}
                          & Acc@0.25          & Acc@0.50          & Acc@0.25 & Acc@0.50                         \\ \midrule

\rowcolor{gray!10} SAM \cite{SAM}              &  52.00                 &  48.10                & 77.60         & 72.05                                                \\ 
EdgeSAM \cite{EdgeSAM}          &  50.87            & 47.17     & 76.98    & 71.63                         \\
\hline\toprule
\end{tabular}

\label{tabel_ablation_sam}
\end{table}

\textbf{Effect of imperfect SAM-based supervision $\mathcal{L}_{\text{SAM}}$.}
SAM's \cite{SAM} masks inevitably contain noise, yet in our framework they act as auxiliary supervision rather than hard constraints. Table \textcolor{red}{2} in the main manuscript shows that adding the SAM-based loss term ($\mathcal{L}_{\text{SAM}}$) consistently improves performance, which indicates that the pseudo-masks provide useful object-level cues despite their imperfections. To further understand the effect of SAM noise, we replace the original SAM with EdgeSAM \cite{EdgeSAM}, a lightweight variant whose predicted masks are noisier. As shown in Table~\ref{tabel_ablation_sam}, this substitution leads to only a slight drop in $\text{Acc@}0.25$ and $\text{Acc@}0.50$ for both Ground$\rightarrow$Satellite and Drone$\rightarrow$Satellite settings, and the performance remains clearly better than training without $\mathcal{L}_{\text{SAM}}$. These results suggest that our model is robust to the noise in SAM masks and can effectively exploit their coarse object extents while ignoring fine-grained errors. Overall, although SAM-based pseudo-masks are imperfect, they still provide beneficial object-level supervision and bring consistent gains to ReCOT.

\subsection{More Comparison Results.}

\begin{table}[!htb]
\caption{Comparisons between DetGeo \cite{DetGeo}, DetGeo using Swin-t \cite{swin} (DetGeo\textsuperscript{*}), VAGeo \cite{VaGeo}, TROGeo \cite{TROGeo}, and our methods in terms of $\text{Acc@}0.25(\%)\uparrow$ and $\text{Acc@}0.50(\%)\uparrow$ on the \emph{test set} of CVOGL dataset. ``$\dagger$" indicates that GeoFormer \cite{GeoFormer} adopts Swin-b as the backbone.}
\centering
\scriptsize
\setlength{\tabcolsep}{0.5mm} 
\renewcommand\arraystretch{1.2} 

\begin{tabular}{{c|cc|cc}}
\hline
\toprule
\multirow{2}{*}{Method} & \multicolumn{2}{c|}{Ground$\rightarrow$Satellite} & \multicolumn{2}{c}{Drone$\rightarrow$Satellite}     \\ \cline{2-5}
                          & Acc@0.25          & Acc@0.50          & Acc@0.25 & Acc@0.50                         \\ \midrule
DetGeo \cite{DetGeo}                 & 45.43            & 42.24             & 61.97    & 57.66                        \\
DetGeo\textsuperscript{*}                  & 46.01             & 41.44             & 66.60    & 56.11                           \\ \midrule
GeoFormer\textsuperscript{$\dagger$} \cite{GeoFormer}                  & 49.54 & 45.32 & 73.79 & 68.96                           \\ \midrule
VAGeo \cite{VaGeo}                  & 48.21             & 45.22             & 66.19    & 61.87                           \\
TROGeo \cite{TROGeo}     & 51.08 & 46.56  & \underline{76.16} & \underline{68.96} \\ \midrule

\rowcolor{gray!10} ReCOT (Ours)              & \textbf{52.00} & \textbf{48.10}                 & \textbf{77.60} & \textbf{72.05}    \\
 \hline
  \toprule
\end{tabular}

\label{tabel_comparison_backbone}

\end{table}

\textbf{Quantitative Results.} As shown in Table \ref{tabel_comparison_backbone}, we replace the backbone of DetGeo~\cite{DetGeo} with Swin-t~\cite{swin} for a more comprehensive and fair comparison. While the upgraded DetGeo\textsuperscript{*} shows slight improvements on Acc@0.25, it still lags behind our ReCOT by a large margin across all evaluation metrics. Moreover, the backbone replacement does not lead to consistent gains, as DetGeo\textsuperscript{*} fails to outperform other recent CVOGL approaches \cite{VaGeo, OCGNet, TROGeo} in Table \ref{tabel_comparison_backbone}. Besides, although GeoFormer \cite{GeoFormer} adopts a larger backbone (Swin-b), it still fails to outperform our ReCOT and the recent CVOGL approach \cite{TROGeo}. These results demonstrate that the key factor in CVOGL does not lie in the backbone. In contrast, our proposed ReCOT achieves superior performance through a more effective and task-aligned framework, highlighting the importance of architectural innovations for CVOGL.

\begin{figure}[!htb]
  \centering
  \includegraphics[width=1.0\linewidth]{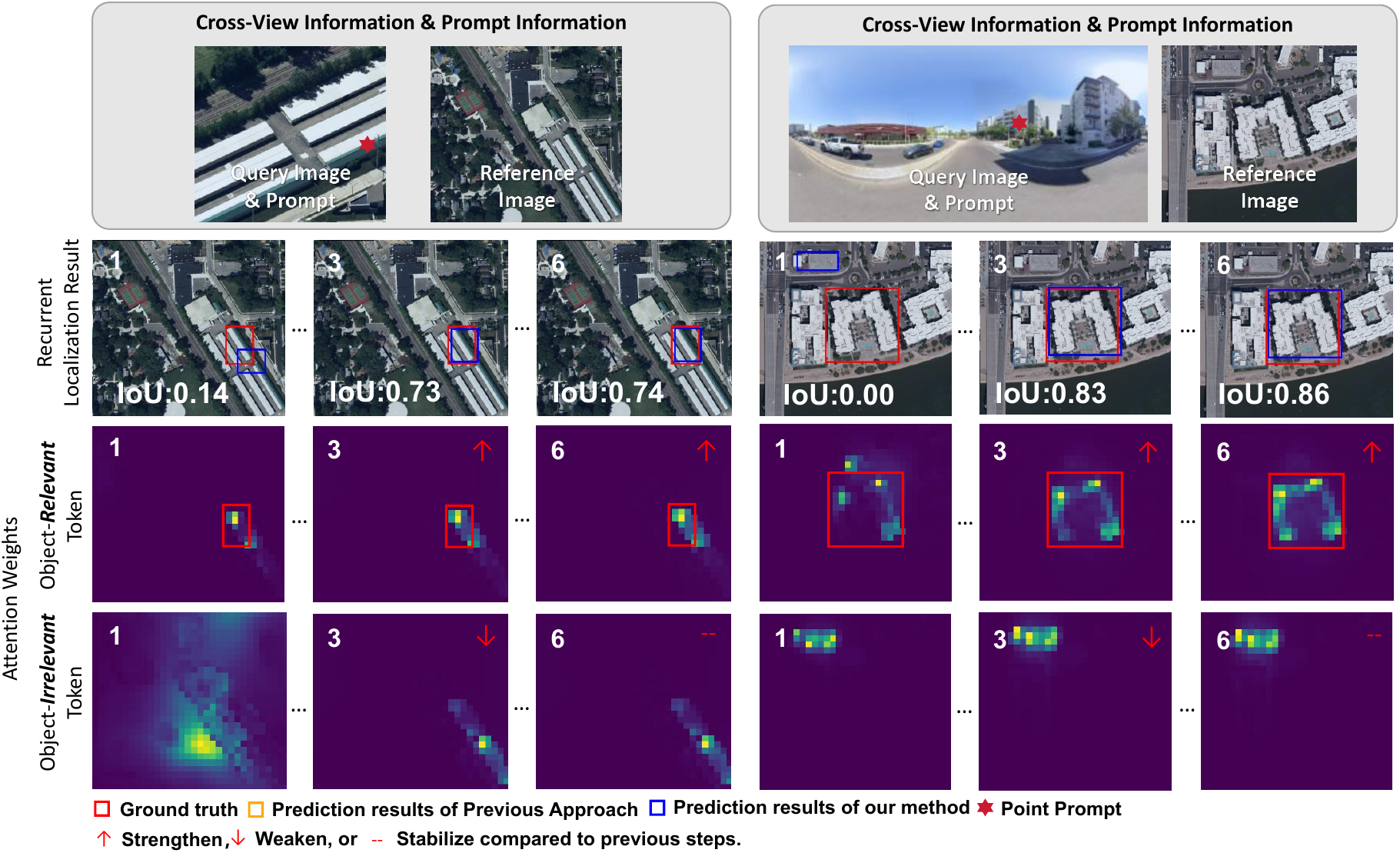}
   \caption{Visualizations of how our ReCOT works. The object-relevant tokens progressively focus and strengthen around the indicated object, while object-irrelevant tokens weaken and stabilize to background patterns.}
   \label{fig: motivation2}
   
\end{figure}

\textbf{Qualitative Results.} Fig. \ref{fig: motivation2} provides more visualization results of the cross attention weights between the token and reference feature during recurrent process. As the recurrent steps progress, object-relevant tokens gradually focus and strengthen around the expected object, enabling the bounding box to converge toward the correct location. In contrast, object-irrelevant tokens weaken and gradually stabilize over iterations. This behavior highlights the ability of ReCOT to dynamically disentangle object-focused information from irrelevant features, which is a key factor driving the success of our recurrent localization strategy.

\subsection{More Discussions}
\textbf{Detailed Distinction from Previous CVOGL Approaches.}
It is important to clarify that our recurrent localization mechanism is fundamentally different from simply stacking more layers in a one-shot architecture. Existing CVOGL methods \cite{DetGeo, OCGNet, VaGeo, TROGeo, AttenGeo, HALO} adopt convolution-based detection heads without explicit iterative correction. Meanwhile, GeoFormer~\cite{GeoFormer} introduces a stacked transformer decoder with independently parameterized layers, which increases depth but does not perform structured state revision across steps. In contrast, ReCOT formulates localization as an unrolled recurrent process with shared parameters across refinement steps. This weight sharing enables iterative state revision within a fixed parameter space, rather than learning a deeper feed-forward mapping as in stacked transformer decoders. Notably, despite using significantly fewer parameters, ReCOT consistently outperforms GeoFormer in our comparisons (Table \textcolor{red}{4} in the main manuscript), suggesting that the performance gains arise from structured recurrent refinement rather than increased model capacity or depth. Moreover, when the proposed token-level supervision is removed, the model degenerates into a DETR-like structure without explicit semantic accumulation, and performance drops accordingly (Table \textcolor{red}{2} in the main manuscript). These observations collectively indicate that the improvements of ReCOT stem from the recurrent refinement mechanism itself, rather than architectural scaling.

\subsection{Limitations and Future Work}

\begin{figure}[!htb]
  \centering
  \includegraphics[width=0.7\linewidth]{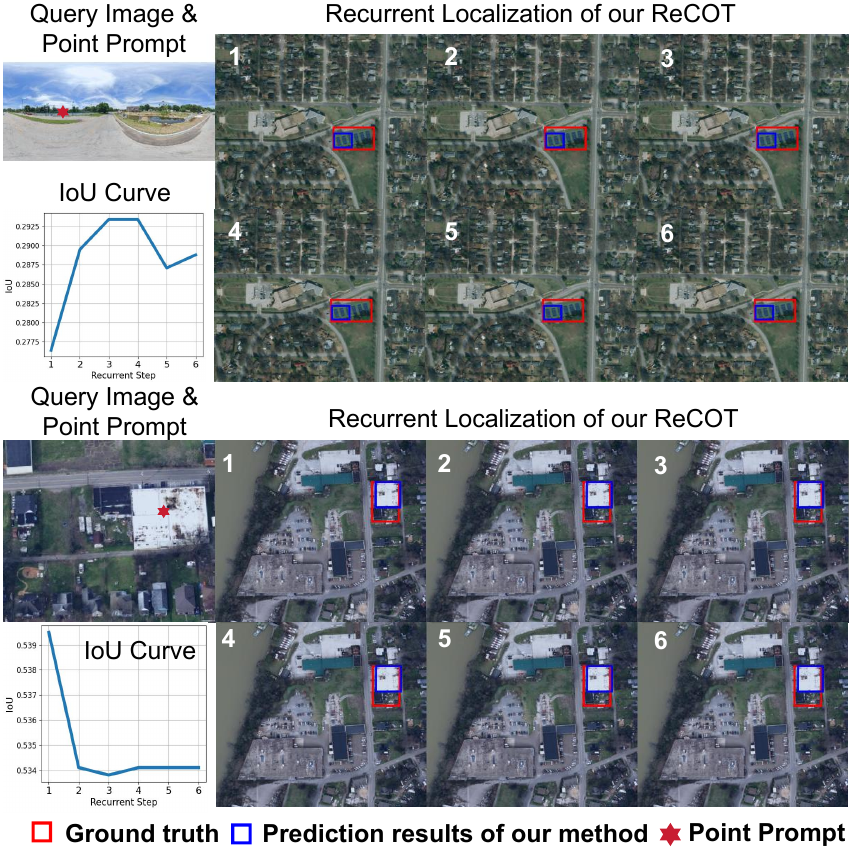}
   \caption{Examples of failure cases. Please refer to the zoomed-in view for better visualization.}
   \label{fig: failure}
    
\end{figure}
As shown in Fig. \ref{fig: failure}, some failure cases of ReCOT are caused by ambiguous or imprecise point prompts, which often highlight only a part of the object rather than its entirety. This ambiguity misguides the token-driven recurrent refinement process, leading to suboptimal localization results. In the future, we will incorporate multi-modal prompts (\emph{e.g}., textual descriptions or bounding boxes) to provide richer and more accurate prompt semantics. Such multi-modal guidance could reduce ambiguity and further improve the robustness and precision of recurrent localization.

\bibliographystyle{splncs04}
\bibliography{main}
\end{document}